# Challenges for cognitive decoding using deep learning methods

Armin W. Thomas [★,◊,*], Christopher Ré [■], and Russell A. Poldrack [★,◊]

★ Stanford Data Science, Stanford University, Stanford, CA, USA
◊ Department of Psychology, Stanford University, Stanford, CA, USA
■ Department of Computer Science, Stanford University, Stanford, CA, USA
*Correspondence: athms@stanford.edu (A. W. Thomas)

In cognitive decoding, researchers aim to characterize a brain region's representations by identifying the cognitive states (e.g., accepting/rejecting a gamble) that can be identified from the region's activity. Deep learning (DL) methods are highly promising for cognitive decoding, with their unmatched ability to learn versatile representations of complex data. Yet, their widespread application in cognitive decoding is hindered by their general lack of interpretability as well as difficulties in applying them to small datasets and in ensuring their reproducibility and robustness. We propose to approach these challenges by leveraging recent advances in explainable artificial intelligence and transfer learning, while also providing specific recommendations on how to improve the reproducibility and robustness of DL modeling results.

---

**Glossary:**

**Cognitive state:** An unobservable construct of psychological theory that refers to a particular mental operation or content and is often associated with specific observable behaviors.

**CV:** Computer vision (CV) is an area of artificial intelligence research, which aims to enable computers to derive meaningful information from the visual world and to take actions based on that information.

**DL:** Deep learning (DL) describes a class of representation learning methods, which transform the input data in multiple sequential steps (or layers), each applying stacks of simple, but nonlinear functions.

**fMRI:** Functional magnetic resonance imaging (fMRI) measures brain activity by detecting changes in activity associated with changes in local blood flow.

**NLP:** Natural language processing (NLP) is an area of artificial intelligence research, which aims to enable computers to derive meaningful information from human language and to take actions based on that information.

**Representation:** As used in computer science, a transform of some data in terms of a different set of features.

**XAI:** Explainable artificial intelligence (XAI) represents a class of methods, which aim to make the behavior of DL methods understandable to human observers, for example, by relating the features of some input data to the respective outputs of the model.

---

# The promise of deep learning

Over the last decade, deep learning (**DL**; see Glossary and [1]) methods have revolutionized many areas of research and industry with their ability to learn highly versatile **representations** (or concepts) from complex data. A defining feature of DL methods is that they sequentially apply stacks of many simple, but nonlinear, transforms to their input data, allowing them to gain an increasingly abstracted view of the data. At each level of the transform, new representations of the data are built by the use of representations from preceding layers. The resulting high-level view of the data enables DL methods to capture complex nonlinearities, associate a target signal with highly variable patterns in the data (e.g., when identifying objects in images or transcribing audio recordings), and effectively filter out aspects of the data that are irrelevant to the learning task at hand. A key driver for the empirical success of DL methods is that they are able to autonomously learn these different levels of abstraction from sufficiently large datasets, without the need for extensive data preprocessing or a prior understanding of the mapping between input data and target signal.

This empirical success has recently sparked interest in the application of DL methods to the field of neuroimaging, focused on cognitive decoding [2]. Here, researchers aim to understand the mapping between a set of **cognitive states** (e.g., answering questions about a math problem vs. a prose story) and the underlying brain activity by training some predictive model to identify these states from measured brain activity [3]. At first sight, DL methods seem ideally suited for these types of analyses, as the mapping between cognitive states and brain activity is often a priori unknown, can be highly variable within [4] and between individuals [5], and is subject to spatial and temporal non-linearities [6].

Yet, the application of DL methods to cognitive decoding analyses also raises several new challenges for researchers who are interested in combining methods from both fields, namely, a general lack of interpretability of DL methods, their overall demand for large training datasets, and difficulties in ensuring the reproducibility and robustness of DL modeling results. Here, we outline each of these challenges and present a set of solutions based on related empirical work and recent methodological advances in both functional neuroimaging and machine learning research.

# Opening up the black box

A key challenge for the application of DL models to functional neuroimaging data is the black-box characteristic of DL models, whose highly non-linear nature deeply obscures the relationships between input data and decoding decisions. Thus, even if a DL model accurately decodes a set of cognitive states from functional neuroimaging data, it is not clear which particular features of the data (or combinations thereof) support this decoding. To approach this challenge, functional neuroimaging researchers have begun turning towards research on explainable artificial intelligence (**XAI**; [7-9]), where techniques are being developed that aim to make the behavior of DL models (and other intelligent artificial agents) understandable for human observers.

One line of research within this field seeks to explain the predictions of DL models by relating these predictions to the features of the input data, thus, making the model interpretable for human observers [10]. On a high level, these explanation techniques can be categorized into those that aim to provide either a global or local explanation of the model's learned mapping between input data and predictions [8,9]. Global explanations [11,12] are not tied to a specific data sample and seek to provide insights into the characteristics of the data at large that influence the model's predictions (e.g., by synthesizing the image for which the model is most certain that it depicts a dog). Local explanations [12-29], on the other hand, are specific for a data sample and seek to provide insights into the mapping between the features of the sample and the resulting prediction of the model (e.g., by identifying those pixels of an image that the model views as evidence for the presence of a dog). Further, explanation techniques can be categorized into those that can be applied post-hoc to the predictions of an already trained model [11-26] and those that are integrated into the training procedure, such that the model is self-explaining and designed a priori to provide human-understandable insights into its learned mappings between data and predictions [27-31].

Here, we focus on local explanation approaches. We provide an overview of a set of representative approaches to this type of XAI in Box 1. Of these approaches, sensitivity analysis [12,22], backward decomposition [13-19,21,24], and attention mechanisms [27-29] are currently most prominent in the functional neuroimaging literature on cognitive decoding with DL models [32-40]. Sensitivity analysis seeks to explain the predictions of a DL model by identifying those features of an input sample to which the model's prediction responds most sensitively. Backward decompositions, on the other hand, seek to explain individual predictions by sequentially decomposing these predictions in a backward pass into the contributions of the lower-layer model units to these predictions, until the input space is reached and a contribution for each input feature can be defined. Attention mechanisms, in contrast, are inspired by research in neuroscience on perceptual attention [41] and aim to improve the predictive performance of a DL model by focusing its computations on specific aspects of an input sample (through a set of attention weights). Attention mechanisms are thus an integrated feature of the training procedure and cannot be applied post-hoc to the predictions of trained models.

At first sight, the explanations of these three XAI techniques can be difficult for human observers to distinguish, making it difficult to compare the quality of their explanations. To approach this challenge, researchers in machine learning have started developing methods to quantify the quality of these types of explanations [8,42-47]. One prominent approach is to test the faithfulness of an explanation [8,42,44,46]. An explanation can generally be viewed as being faithful [48] if it accurately captures the model's decision process and thereby identifies those features of the input that are most relevant for the model's prediction. Accordingly, removing these features from the input (e.g., in an occlusion analysis; [8,42,44]) should lead to a meaningful decline of the model's predictive performance.

**Box 1. Exemplary approaches to local explainable artificial intelligence.**
We assume that the analyzed model represents some function $f(\cdot)$, mapping an input $x \in \mathbb{R}^D$ to some output $f(x)$: $f(\cdot): \mathbb{R}^D \rightarrow \mathbb{R}$. The presented local explanation approaches $\eta(\cdot)$ seek to provide insights into this mapping by quantifying the relevance $r_d$ of each input feature $d \in D$ for $f(x)$: $\eta(\cdot): \mathbb{R} \rightarrow \mathbb{R}^D$ (Fig. I).

**Occlusion analysis** [20,24,25]**:** The occlusion analysis is a perturbation analysis, which identifies $r_d$ by occluding $x_d$ in the data and measuring the resulting effect on $f(x)$: $r_d = \Delta f(x) = f(x) - f(x \times o_d)$. Here, $o_d$ indicates an occlusion vector (e.g., $o_d \in [0,1]^D$), and $\times$ the element-wise product.

**Interpretable local surrogate model** [26]**:** A local surrogate model is an interpretable model that is used to explain black-box model predictions by training it to approximate these predictions. In the LIME algorithm [26], $r$ is quantified by approximating $f(x)$ for a specific $x$ with an interpretable model $g(\cdot)$, e.g., a linear model, where $g(x) = w^T x$, and which is trained by the use of a set of perturbed versions $Z$ of $x$ (e.g., through occlusion): $\min_{w} \sum_{z \in Z} \pi_x(z) \, (f(z) - g(z))^2$ . Here, $\pi_x$ represents some similarity function weighting each $z \in Z$ by its similarity to $x$, while $r_d$ is given by the corresponding linear model weight $w_d$: $r_d = w_d$.

**Sensitivity analysis** [12,22]**:** Sensitivity analysis defines $r$ as the locally evaluated partial derivative of $f(x)$: $r = f'(x) = (\frac{\partial f(x)}{\partial x})^2$. Accordingly, relevance is assigned to those features $x_d$ to which $f(x)$ responds most sensitively.

**Backward decomposition** [13-19,21,24]**:** Backward decompositions make specific use of the graph structure of DL models by sequentially decomposing $f(x)$ in a backward pass through the model until the input space is reached. A prominent example is the layer-wise relevance propagation (LRP; [13]) technique: Let $i$ and $j$ be the indices of two model units in two successive layers, $r_j$ the relevance of unit $j$ for $f(x)$, and $r_{i \leftarrow j}$ the part of $r_j$ that is redistributed from $j$ to $i$ during the backward decomposition. To redistribute relevance between layers, several rules have been proposed [49,50], which generally follow from: $r_i = \sum_j \frac{a_i w_{ij}}{\sum_i a_i w_{ij}} r_j$, where $a$ and $w$ represent the input and weights of unit $i$. Importantly, LRP assumes that relevance is conserved between layers, such that $\sum_i r_{i \leftarrow j} = r_j$, $r_i = \sum_j r_{i \leftarrow j}$, and $\sum_d r_d = \ldots = \sum_i r_i = \sum_j r_j = f(x)$.

**Attention mechanism** [27-29]**:** In contrast to the other presented explanation approaches, attention mechanisms are built into and trained with a DL model. Their goal is to help the model focus its computations on specific aspects of $x$ by the use of attention glimpses $\gamma \in \mathbb{R}^D$. These glimpses can be implemented by the use of a separate model, which predicts an attention weight $\alpha_d$ for each $x_d$ [27,28]. These weights are multiplied with the input ($\gamma = \alpha \times x$) and the resulting glimpse is forwarded to the remaining parts of the model (instead of $x$). Attention weights are often also interpreted as an estimate of relevance, such that $r_d = \alpha_d$ [32,42,43].

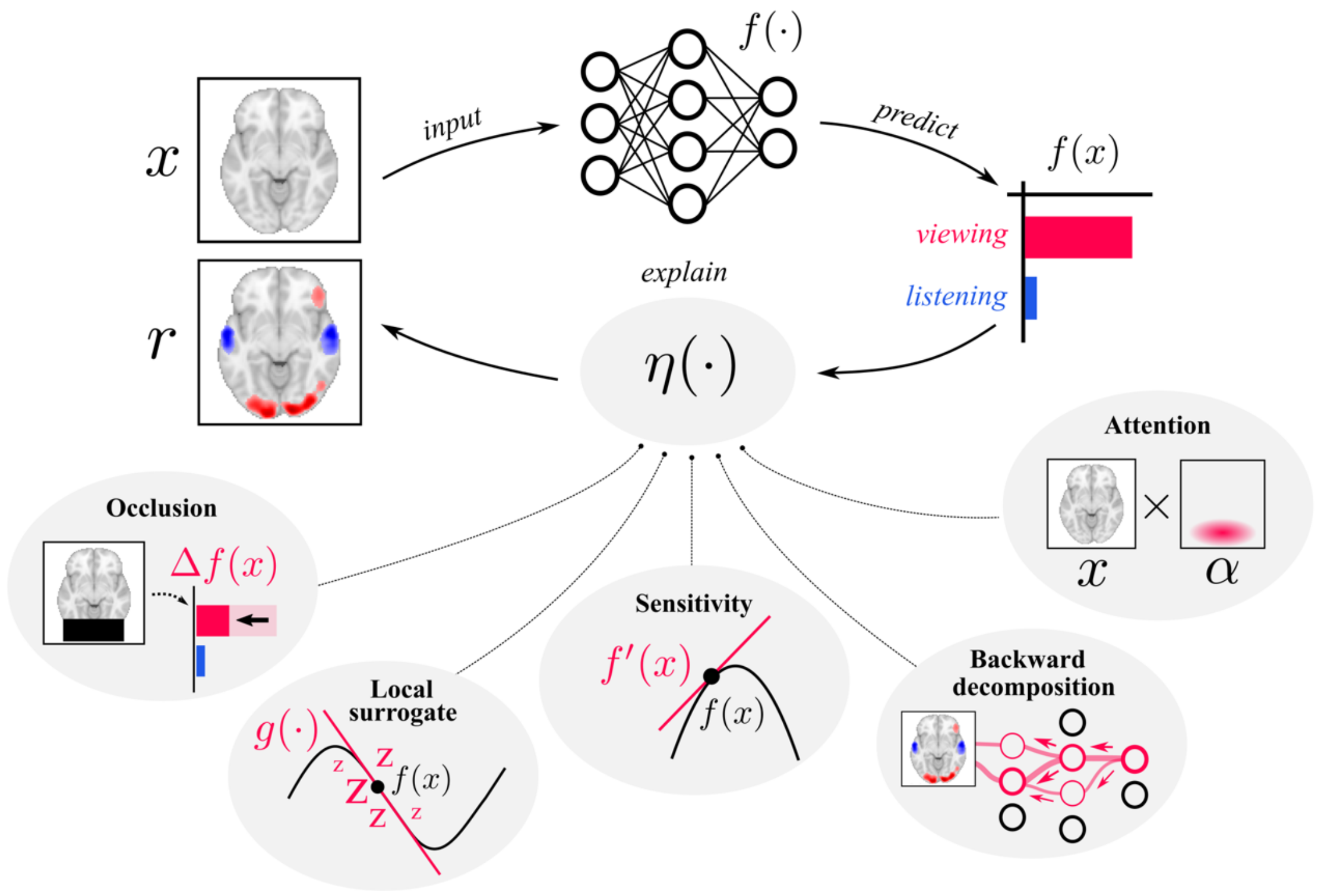

**Figure I.** Exemplary approaches to local explainable artificial intelligence.

---

By the use of this test, researchers in computer vision (**CV**; [44]) and natural language processing (**NLP**; [42,43]) have compared the fidelity of explanations resulting from sensitivity analysis and backward decomposition. This work has shown that backward decompositions generally perform better at identifying those features of the input that are most relevant for model predictions. Intuitively, this makes sense, as backward decompositions seek to directly quantify the contribution of each input feature to a specific model prediction. Sensitivity analysis, in contrast, does not evaluate the prediction itself but its local slope, thus identifying features that make the model more or less certain of its prediction, regardless of their actual contribution to the prediction. Similarly, related empirical work [42,43,46] has demonstrated that standard attention mechanisms do not provide faithful explanations of the predictions of DL models, as they fail to identify features of the input that are relevant for the prediction. Attention mechanisms are designed to improve the predictive performance of a DL model by enhancing certain aspects of the input while fading out others. Their decisions on how to weight each input feature are thus tailored to the subsequent computations of the DL model and not designed to identify the relevance of each feature for the prediction.

While functional neuroimaging researchers have also used the occlusion analysis to analyze cognitive decoding models (in "virtual lesion analyses"; [35,51]), these applications have been mostly limited to linear models and to testing whether specific brain regions, which received large weights in a linear model, are actually necessary for an accurate decoding. For functional neuroimaging data, occlusion analyses generally require a clear prior hypothesis on which features (or brain regions) of the input will be tested (e.g., resulting from other research or other explanation analyses), as randomly dropping out individual feature values will otherwise not account for the strong spatial correlation structure inherent to these data.

Taken together, we thus generally recommend the application of backward decompositions to analyze cognitive decoding decisions of DL models (see Box 2 for specific recommendations).

In addition to selecting an XAI technique, functional neuroimaging researchers interested in understanding the decoding decisions of deep cognitive decoding models are also faced with the task of analyzing the resulting dataset of explanations (which is generally equal in size to the original input data, as one explanation is obtained for each input sample). Many current empirical applications of XAI in cognitive decoding simply average individual explanations or analyze them with standard linear models [32-34,36-38]. These types of linear analyses, however, are limited in their ability to identify any non-linear mappings between brain activity and cognitive states that a DL model might have learned. This problem has been similarly encountered in machine learning research. Here, researchers have instead advocated for the application of automated, non-linear, and data-driven approaches to identifying the learned decision behaviors of DL models from explanation data (e.g., through t-distributed stochastic neighbor embedding; [52]). In functional neuroimaging, current methods from network neuroscience [53] also seem highly promising for analyzing explanation data, by identifying the individual units (e.g., brain regions) of the learned mappings between brain activity and decoded cognitive states as well as the dynamics of their interactions.

---

**Box 2. Recommended XAI approaches for cognitive decoding.**

We generally recommend the application of backward decompositions (see Box 1 and [13-19,21,24]) to analyze the contribution of individual input features to the cognitive decoding decisions of DL models. Below, we provide three specific recommendations for XAI approaches for cognitive decoding:

**LRP** [13]: An overview of LRP is provided in Box 1. While several rules have been proposed to redistribute relevance $r$ between units $i$ and $j$ of two successive layers [13,49,50], recent empirical work has shown that a composite of these rules is best-suited for CV models [49,50]. Specifically, this work suggest combining the LRP-0 rule ($r_i = \sum_j \frac{a_i w_{ij}}{\sum_i a_i w_{ij}} r_j$, where $a$ and $w$ represent the input and weights of unit $i$) for layers closer to the output, with the LRP-$\epsilon$ rule ($r_i = \sum_j \frac{a_i w_{ij}}{\epsilon + \sum_i a_i w_{ij}} r_j$ with $1e^{-4} \leq \epsilon < 1$) for middle layers, and the LRP-$\gamma$ rule ($r_i = \sum_j \frac{a_i(w_{ij} + \gamma w_{ij}^{+})}{\sum_i a_i(w_{ij} + \gamma w_{ij}^{+})} r_j$, where $\gamma$ controls positive contributions and is generally $0 < \gamma$) for layers closer to the input. A TensorFlow implementation of LRP is provided by iNNvestigate [54], while Captum [55] provides PyTorch implementation.

**Deep learning important features (DeepLIFT)** [16]: DeepLIFT seeks to explain the contribution of model unit $i$ to prediction $f(x)$ by comparing the unit's current activation $a_i$ to a reference activation $a_i^0$, which results from a reference input $x^0$[16]. The reference input $x^0$ should be neutral and preserve only those properties of the input that are not considered relevant to the decoding problem while removing properties that are considered relevant (e.g., for MNIST, an all-zero reference is recommended, as this is the MNIST background). For functional neuroimaging data, we thus recommend an all-zero reference, a reference that adds noise to the input as well as a reference involving samples from other decoding classes (e.g., their average). We further recommend using multiple references to increase robustness towards specific reference choices [23]. A TensorFlow implementation of DeepLIFT can be found at github.com/kundajelab/deeplift and github.com/slundberg/shap, while Captum [55] provides a PyTorch implementation.

**Integrated gradients (IG)** [15]: The integrated gradients (IG) technique is closely related to both LRP and DeepLIFT [56]. IG is applicable to any differentiable model and defines relevance $r_d$ of input feature $d$ for prediction $f(x)$ by integrating the gradient $f'(x)$ along some trajectory in the input space connecting a reference point $x^0$ to the current input $x$: $r_d = (x_d - x_d^0) \int_{\alpha=0}^{1} \frac{\delta f(x^0 + \alpha(x - x^0))}{\delta x_d} d\alpha$. Similar to DeepLIFT, the authors recommend an all-zero reference or the addition of noise to the input [14,57], while an average over multiple references is also possible [57]. A tutorial on how to use IG in TensorFlow can be found at tensorflow.org/tutorials/interpretability/integrated_gradients, while Captum [55] provides a PyTorch implementation.

---

# Leveraging public data

A second major challenge for DL models in functional neuroimaging research is the high dimensionality and low sample size of conventional functional neuroimaging datasets. A typical functional Magnetic Resonance Imaging (**fMRI**) dataset contains a few hundred volumes for each of tens to hundreds of individuals, while each volume contains several hundred thousand voxels (or dimensions). Current state-of-the-art DL models, in contrast, can easily contain many hundred million parameters [58,59], while recent language models have pushed this boundary even further with many billion parameters [60-62]. In most cases, DL models thus contain many more trainable parameters than there are samples in their training data. While this vast overparameterization represents a key element to the empirical success of DL models, by enabling them to find near-perfect solutions for most standard learning tasks [63,64] and to generalize well between datasets [62,65,66], it also represents one of the biggest challenges for their application in fields where data are scarce, as the performance of DL models is strongly dependent on the amount of available training data [61,67].

To approach this challenge, researchers have developed many methods that aim to improve the performance of DL methods in smaller datasets (e.g., [68-71]). One of these methods, with strong empirical success, is transfer learning [71]. The goal of transfer learning is to leverage the knowledge about a mapping between input data and a target variable that can be learned from one dataset (i.e., the source or upstream domain) to subsequently improve the learning of a similar mapping in another dataset of a related domain (i.e., the target or downstream domain). Knowledge is typically transferred in the form of the parameters that a model has learned in the source domain and that are then used to initialize the model (or parts of the model) when beginning learning in the target domain. Transfer learning has been especially successful in CV and NLP, where large publicly available datasets exist (e.g., [72,73] and http://www.commoncrawl.org). Here, DL models are first pre-trained on these large datasets (e.g., to classify objects in images) and subsequently fine-tuned on smaller datasets of a related target domain (e.g., to classify brain tumors in medical imaging; [74]). Computationally, pre-training can aid subsequent optimizations by placing the parameters near a local minimum of the loss function [75] and by acting as a regularizer [76]**.** Pre-trained models generally exhibit faster learning and higher predictive accuracy, while also requiring less training data when compared to models that are trained from scratch [61,65,66,71,75-78]. The benefits of pre-training can diminish, however, with increasing size of the target dataset [61] and as the overall differences between source and target learning task and/or domain increase [79,80].

Over recent years, the field of functional neuroimaging has experienced a similar increase in the availability of public datasets, which are provided by large neuroimaging initiatives as well as individual researchers [81]. In addition, several efforts have been made to standardize the organization (e.g., [82,83]) and preprocessing (e.g., [84]) of functional neuroimaging data. These developments have paved the way for the field of functional neuroimaging to enter a big data era, allowing for the application of transfer learning.

First empirical evidence indicates that transfer learning between individuals [85-87], experiment tasks [36, 37, 39,88], and datasets [89-92] is possible and that pre-training generally improves the performance of DL models when applied to conventional fMRI datasets [36,85,86,88,89,91]. Most of this work has utilized traditional supervised learning techniques during pre-training, by assigning a cognitive state to each sample in the data and training a decoding model to identify these states from the data (see Box 3). While this is a fruitful approach to decoding analyses within individual neuroimaging datasets, it is often difficult to extend to analyses across many datasets. In spite of several attempts (e.g., [93,94]), functional neuroimaging research has yet to widely adopt standardized definitions of cognitive states. Without this type of standardization, it is often unclear whether two experiments from two separate laboratories elicit the same or different sets of cognitive states. Imagine the following experiments: In the first, participants read aloud a sequence of sentences and are then asked to repeat the last word of each sentence [95]. In the second, participants first hear a sequence of letters and digits and are then asked to report back the letters and digits in alphabetical and numerical order respectively (the letter–number sequencing task; [96]). While both experiments label the associated cognitive state as “working memory”, one could argue that the experiments in fact elicit two distinct cognitive states, as one solely requires temporarily storing information while the other also requires active manipulation of this information.

This problem of imprecisely labeled data has been similarly encountered in machine learning research. Here, researchers have developed weakly supervised learning techniques, which aim to train DL models with noisy or incomplete data labels [97]. One prominent weak supervision approach is data programming (see Box 3 and [98,99]), which seeks to alleviate the cost of manually labeling large datasets by the use of simple labeling functions. These functions automatically label subsets of the data by implementing simple domain heuristics of subject matter experts (e.g., label a YouTube text comment as Spam if it contains a URL or the words "check this out"). The generated labels are then used to train DL models in a supervised manner. Importantly, data programming frameworks account for noise and conflicts that can arise from the automatic labeling by representing the labeling process as a generative model [98,99]. Recent empirical work has demonstrated that this type of weak supervision can be successfully used for the classification of unlabeled medical imaging data (e.g., radiography or computer tomography data; [100]), by designing labeling functions that extract labels from the accompanying medical text reports. A similar approach could be fruitful to generate standardized labels of cognitive states (e.g., according to the Cognitive Atlas; [94]) by applying these types of automatic labeling functions to the accompanying publication texts (e.g., label an fMRI scan as "visual perception" if the publication text contains the words "viewed" or "viewing" in the Methods section).

An alternative approach, with strong recent empirical success, is unsupervised learning (see Box 3 and [101,102]). Unsupervised learning does not consider any labeling and instead trains models to autonomously learn meaningful representations of the data. These learned representations can then be used to improve learning in a related target domain. Two prominent examples of unsupervised learning are contrastive and generative learning [103]. Both learn a representation of the data by training an encoder model to project the data into a lower-dimensional representation, which preserves relevant information. In contrastive learning [104], the encoder model is trained by the use of an additional discriminator model, which aims to determine the similarity of a pair of data samples based on their projection through the encoder model. During training, positive pairs of samples are created by randomly augmenting the same data sample twice (e.g., different views of the same image), while negative pairs are created by augmenting two different data samples. Generative learning [105], in contrast, trains the encoder model by the use of an additional decoder model, which aims to reconstruct the original data sample from the lower-dimensional representation of the encoder model.

---

**Box 3. Approaches to pre-training.**

Transfer learning aims to improve the performance of model $f(\cdot)$ in a target learning task $T_T$ in a target domain $D_T$ by leveraging knowledge that can be learned by *pre-training* $f(\cdot)$ in a related source learning task $T_S$ and source domain $D_S$ [71]. Knowledge is generally transferred through a set of model weights $W$ that $f(\cdot)$ has learned during pre-training. A domain $D$ is defined by a feature space $X$ with samples $x \in \mathbb{R}^N$ whose $N$ feature values are characterised by some probability distribution $P(X)$. Given a domain, a learning task $T$ consists of learning some mapping between a label space $Y$ and the associated feature space $X$, by training $f(\cdot)$ to accurately predict the target values $y \in Y$ assigned to the samples $x \in \boldsymbol{X}$ in a dataset $A = \{(x_i, y_i), \dots, (x_m, y_m)\}$, such that $f(x_i) \approx y_i$. Here, we describe three learning tasks (Fig. I), which enable $f(\cdot)$ to learn in a source domain, when the target values $y$ of a source dataset $A_S$ are either fully accessible, noisy or incomplete, or not accessible.

**Supervised learning** [106]: In supervised learning, the target values $y \in A_S$ are fully accessible and can be used to train $f(\cdot)$, for example, by minimizing the distance between $f(x)$ and $y$: $\min_W \sum_{\{x_i, y_i\} \in A_S} (y_i - f(x_i))^2$.

**Weakly supervised learning** [97]: In weakly supervised learning, the target values $y \in A_S$ are noisy or incomplete. A prominent example of weak supervision is data programming [98], where noisy target values $\hat{y}$ are generated for an unlabeled source dataset by the use of user-specified labeling functions. These labeling functions implement domain heuristics of subject matter experts (e.g., label a chest radiograph as "abnormal" if the corresponding medical text report contains a word with the prefix ''pneumo''; [100]). The generated target values are then used to train $f(\cdot)$ in a supervised way.

**Unsupervised learning** [101,102]: In unsupervised learning, the target values $y \in A_S$ are not accessible. Instead, a new learning task is devised, which requires $f(\cdot)$ to independently learn a representation of the data in the source domain. Two prominent unsupervised learning strategies are contrastive and generative learning. Both treat $f(\cdot)$ as an encoder model, which is trained to project the samples $x \in A_S$ into a lower-dimensional representation that preserves relevant information: $f(\cdot): \mathbb{R}^N \rightarrow \mathbb{R}^L$ (where $L < N$). In contrastive learning [104], $f(\cdot)$ is trained by the use of an additional discriminator model $d(\cdot): \mathbb{R}^L \rightarrow \mathbb{R}$, which learns to determine the similarity of a pair of data samples based on the encoder's projection. During training, a set of augmented versions $\{\hat{x}\}$ of the data samples $x \in A_S$ is created and the discriminator's task is to identify pairs $\{\hat{x}_k, \hat{x}_j\}$ that result from the same sample $x_i$. In generative learning [105], $f(\cdot)$ is trained by the use of an additional decoder model $d(\cdot): \mathbb{R}^L \rightarrow \mathbb{R}^N$ , which aims to reconstruct the original data sample from the encoder's projection: $d(f(x_i)) \approx x_i$.

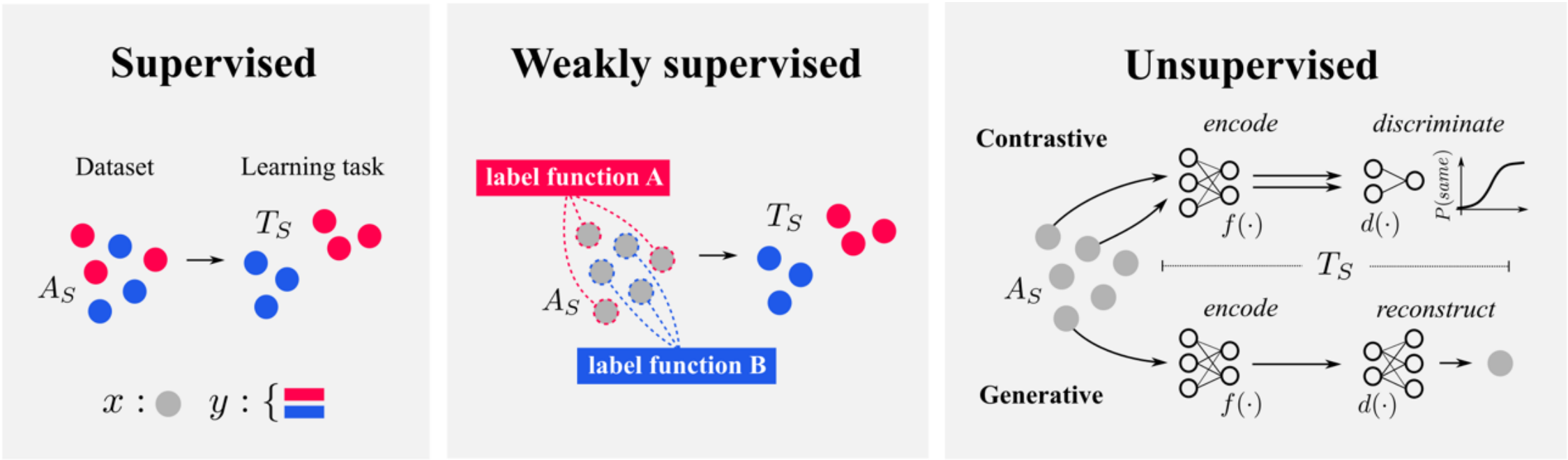

**Figure I.** Illustration of three exemplary source learning tasks ($T_S$), given some source dataset ($A_S$).

---

# Ensuring reproducibility and robustness

Recent work in functional neuroimaging has exposed the high flexibility of its standard analysis workflows, leading to substantial variability in results and scientific conclusions [107]. In light of these issues, various efforts have been made to improve the standardization and reproducibility of functional neuroimaging analyses (e.g., [82,84]). DL research is currently facing similar concerns, with model performances that are often hard to reproduce [108-111] and not robust to the diversity of real-word data [79,112-122]. Functional neuroimaging researchers who are interested in applying DL methods to cognitive decoding analyses are thus faced with additional challenges for the reproducibility and robustness of their work, which arise at the intersection of both fields.

Methodological progress in DL research is often driven by a hunt for state-of-the-art performances in benchmarks (see http://www.paperswithcode.com/sota), that is, by whether a new methodology outperforms existing ones in pre-defined test datasets. While this approach has helped the field of DL to evolve fast and quickly develop accurate models, it has also established a research culture that often sacrifices scientific rigor for maximal performance metrics [123,124], not unlike the “p-hacking” phenomenon in null hypothesis testing [125].

A central argument for predefined test datasets is that all models should be compared on the same grounds (i.e., the same sets of training and testing samples). Yet, these types of point estimates are often insufficient to determine whether a model actually outperforms others in new data. Recent empirical work has demonstrated, for example, that the convergence of DL models and thereby their final performance in a test dataset is dependent on many non-deterministic factors of the training, such as random weight initializations and random shufflings or augmentations of the data during training [110,111,126], as well as the specific choices for hyper-parameters, such as the specification of individual model layers and optimization algorithm [111,127,128]. In some cases, researchers can thus achieve state-of-the-art performance simply by investing large computational budgets into tuning these types of factors for a specific test dataset. Consequently, many of the currently reported DL benchmarks are built on top of massive computational budgets and are often difficult to reproduce by other researchers without full access to the original code and computing environments [110,111,126,129]. Recent empirical findings further suggest that the comparisons performed on several of these benchmarks lack the statistical power required to accurately determine the reported improvements in model performance [130], a problem similarly evident in neuroimaging research [131].

In addition to these challenges for the reproducibility of benchmark model performances, recent findings have also demonstrated that the resulting highly tuned models often lack basic robustness towards slight distributional shifts [112,113,115-117,132] or corruptions [121,133] of the data, while model performances can also vary widely across the different subpopulations of a dataset (stratified, for example, by people's age, race or gender; [134]) or be based in spurious shortcuts that the models have learned from a training dataset but which do not generalize well to other scenarios [52,118].

For these reasons, researchers have started advocating for more comprehensive and standardized reporting of the training history of DL models [135], more extensive evaluation procedures [112-115,136], as well as an increased scientific rigor in DL research [123]. In the following, we briefly outline several recommendations resulting from this work, which aim to improve the reproducibility and robustness of model performances (for a brief overview, see Box 4).

Most DL training pipelines are too complex to allow for a comprehensive evaluation of all possible choices of hyper-parameters and other non-deterministic factors of the training. However, evaluating only a specific instance of these choices does not give a reliable estimate of a model's expected performance in new data. Instead, recent empirical work suggests to randomize as many of these choices as possible over multiple training runs, given the computational budget at hand [136]. This allows to better account for the variance in model performance associated with these choices and thus to obtain a more accurate estimate of the model's expected performance (see [126,135,136]).

In addition, researchers have advocated for the use of multiple random splits of the data into training, validation, and test datasets to evaluate model performances (e.g., with bootstrapping or cross-validation; [109,136]). A single, predefined test dataset generally contains limited information about the whole underlying data distribution and is thus also limited in its ability to provide an accurate estimate of the model's expected performance in new data. Multiple random splits, in contrast, better account for the variance in model performance associated with different splits of the data, allowing to reduce the error in a model's expected performance.

Further, to ensure that the chosen combination of statistical comparison method and test dataset size provides sufficient statistical power to accurately determine the studied difference in model performance, researchers have recently suggested the use of simple simulation studies, by first identifying and estimating the required quantities of the statistical testing procedure (e.g., McNemar's test for paired data requires the models' probabilities of making a correct prediction as well as their agreement rate), and subsequently using these estimates to simulate model comparisons at different dataset sizes [130]. In addition to ensuring that the chosen performance evaluation procedure does not lack statistical power, recent work in neuroimaging also suggests to control for multiple sequential model comparisons, as multiple sequential hypothesis tests (e.g., performance comparisons) on the same dataset can inflate false positive rates [137].

Next to choosing appropriate means of evaluating model performances in a given dataset, it is essential to ensure that the resulting performances are robust towards real-world data (e.g., neuroimaging data from different individuals, tasks, and laboratories). Model performances can falter, for example, when encountering slight distributional shifts or corruptions of the data, as recently demonstrated by researchers who confronted benchmark DL models with datasets that mimicked the original benchmarks but included new samples from the original data sources (thus introducing distributional shifts; [79]) or by corrupting the inputs with simple noise [121,133]. To strengthen robustness towards these types of scenarios, empirical work suggests the use of random data augmentations during training, for example, by randomly flipping images from left-to-right [138] or by randomly occluding [139] or mixing [140,141] inputs. It can also be beneficial to mix different data augmentation strategies [142] or to automatically learn these strategies from the data [143].

Similarly, DL model performances have been shown to often vary highly across the different, often unrecognized, subpopulations of a dataset (a phenomenon known as "hidden stratification"; [119,120]). For example, a DL model trained to distinguish benign and malignant tumors from medical imaging data can perform well on average, while consistently misclassifying an aggressive but rare form of cancer. To detect and measure these types of hidden stratification, recent work suggests three approaches [119]: In schema completion, domain experts define and provide a comprehensive set of subclasses for the test dataset (e.g., a more comprehensive labeling of the cognitive states at different time points of the experiment), which can then be used to evaluate the performance of models trained on the more coarsely-labeled training data. In error auditing, domain experts retrospectively inspect instances of false model predictions to identify possible hidden subpopulations (e.g., in combination with XAI techniques; see Box 1). In automatic algorithmic approaches, dedicated algorithms independently search for hidden subpopulations in the data and/or model predictions (e.g., by the use of clustering techniques). Once hidden stratification is detected, modern DL techniques can be used to improve model performances on subpopulations (e.g., [144]).

Lastly, it was demonstrated that DL models can be susceptible to learning spurious shortcuts that allow them to perform well in a given training dataset but which do not generalize well to other scenarios [52,118,122]. Researchers found, for example, that a pneumonia detection model trained with medical imaging data can learn to perform well on average solely by learning to identify hospital-specific artifacts in the medical images in addition to learning the hospitals' pneumonia prevalence rates [145]. To identify these types of shortcuts, researchers recommend to evaluate model performances on out-of-distribution data, for example, on data from different sources (e.g., neuroimaging data from different laboratories and individuals), and to inspect instances of the data whenever out-of-distribution error rates are high relative to in-distribution errors (e.g., with visual inspection and/or the application of XAI techniques; see Box 1).

**Box 4. Recommendations to improve the reproducibility and robustness of DL modelling results in cognitive decoding.**

The performances of DL models in benchmarks are often difficult to reproduce by other researchers or in new data, as the convergence of DL models (and thereby their final performance) is strongly dependent on many non-deterministic aspects of the training [110,111,135,136]. Further, the resulting highly tuned benchmark performances are often not robust towards the diversity of real-world data [79,112-116]. Below, we provide a set of general recommendations to improve the reproducibility and robustness of DL model performances in cognitive decoding analyses:

◊ Use multiple training runs to estimate a model's expected performance, given the computational budget at hand (for methodological suggestions on how to estimate a model's expected performance (or to perform model comparisons) based on multiple training runs, see [126,135,136]).

◊ Apply random augmentations to the data during training, such as random occlusions [139] and combinations [140,141] of the inputs.

◊ For each training run, randomize as many aspects of the training pipeline as possible (including random seeds, random weight initializations, and random shufflings of the training data) and use a random split of the data into training, validation, and test datasets (e.g., by the use of bootstrapping or cross-validation; [109,136]).

◊ If model comparisons are performed, ensure that the chosen combination of statistical comparison procedure and test dataset size has enough statistical power to accurately determine the studied differences in model performance (e.g., by the use of simple simulation studies; see [130]).

◊ Whenever possible, evaluate model performances on out-of-distribution data (e.g., by using neuroimaging data from different laboratories and individuals) and test for hidden stratification [119,120] by the use of schema completion, error auditing or automatic algorithmic approaches (for methodological details, see [119]).

◊ Finally, publicly share any used data, analysis code, and computing environment (e.g., by the use of containerization with Docker or Singularity) in a dedicated repository (e.g., Open Science Framework; [146]).

---

# Concluding remarks

DL methods have experienced great success in research and industry and have had major impacts on society [1]. This success has triggered interest in their application to the field of cognitive decoding, where researchers aim to characterize the representations of different brain regions by identifying the set of cognitive states that can be accurately decoded (or identified) from the activity of these regions. DL methods hold a high promise to revolutionize cognitive decoding analyses with their unmatched ability to learn versatile representations of complex data. Yet, fully leveraging the potential of DL methods in cognitive decoding is currently hindered by three main challenges, which result from a general lack of interpretability of DL methods as well as difficulties in applying them to small datasets and in ensuring their reproducibility and robustness.

Here, we have provided a detailed discussion of these three challenges and proposed a set of solutions that are informed by recent advances in functional neuroimaging and machine learning research. In sum, we recommend that researchers utilize XAI techniques to identify the mapping between cognitive states and brain activity that a DL model has learned (Box 1-2), improve the performance of DL methods in conventional neuroimaging datasets by pre-training these models on public neuroimaging data (Box 3), and follow a set of specific recommendations to improve the reproducibility and robustness of DL model performances (Box 4). We hope that researchers will take inspiration from our discussion and explore the many open research questions that remain on the path to determining whether DL methods can live up to their promise for cognitive decoding.

# Acknowledgments

Armin W. Thomas is supported by Stanford Data Science through the Ram and Vijay Shriram Data Science Fellowship. Russell A. Poldrack is supported by the National Science Foundation under Grant No. OAC-1760950. Christopher Ré gratefully acknowledges the support of NIH under No. U54EB020405 (Mobilize), NSF under Nos. CCF1763315 (Beyond Sparsity), CCF1563078 (Volume to Velocity), and 1937301 (RTML); ONR under No. N000141712266 (Unifying Weak Supervision); the Moore Foundation, NXP, Xilinx, LETI-CEA, Intel, IBM, Microsoft, NEC, Toshiba, TSMC, ARM, Hitachi, BASF, Accenture, Ericsson, Qualcomm, Analog Devices, the Okawa Foundation, American Family Insurance, Google Cloud, Salesforce, Total, the HAI-AWS Cloud Credits for Research program, Stanford Data Science, and members of the Stanford DAWN project: Facebook, Google, and VMWare. The Mobilize Center is a Biomedical Technology Resource Center, funded by the NIH National Institute of Biomedical Imaging and Bioengineering through Grant P41EB027060. The U.S. Government is authorized to reproduce and distribute reprints for Governmental purposes notwithstanding any copyright notation thereon. Any opinions, findings, and conclusions or recommendations expressed in this material are those of the authors and do not necessarily reflect the views, policies, or endorsements, either expressed or implied, of NIH, ONR, or the U.S. Government.